\documentclass{article}

\usepackage[ruled,vlined]{algorithm2e}
\usepackage{amsmath}
\usepackage{amssymb}
\usepackage[colorlinks,linkcolor=blue]{hyperref}

\usepackage{microtype}
\usepackage{graphicx}
\usepackage{subfigure}
\usepackage{booktabs} 

\usepackage{hyperref}

\newtheorem{theorem}{Theorem}
\newtheorem{lemma}{Lemma}

\usepackage[accepted]{icml2021}

\icmltitlerunning{Weighted Neural Tangent Kernel}

\begin{document}

\twocolumn[
\icmltitle{Weighted Neural Tangent Kernel: A Generalized and Improved Network-Induced Kernel}




\icmlsetsymbol{equal}{*}

\begin{icmlauthorlist}
\icmlauthor{Lei Tan}{sjtu}
\icmlauthor{Shutong Wu}{sjtu}
\icmlauthor{Xiaolin Huang}{sjtu}
\end{icmlauthorlist}

\icmlaffiliation{sjtu}{Department of Automation, Shanghai Jiao Tong University, Shanghai, China}

\icmlcorrespondingauthor{Lei Tan}{lei.tan19@outlook.com}
\icmlcorrespondingauthor{Shutong Wu}{wust20@sjtu.edu.cn}
\icmlcorrespondingauthor{Xiaolin Huang}{xiaolinhuang@sjtu.edu.cn}

\icmlkeywords{Machine Learning, ICML}

\vskip 0.3in

]
\footnote{Shanghai Jiao Tong University, Shanghai, China. Correspondence to: Lei Tan $<$ lei.tan19@outlook.com $>$, Shutong Wu $<$ wust20@sjtu.edu.cn $>$, Xiaolin Huang $<$ xiaolinhuang@sjtu.edu.cn $>$.}




\begin{abstract}
The Neural Tangent Kernel (NTK) has recently attracted intense study, as it describes the evolution of an over-parameterized Neural Network (NN) trained by gradient descent. 
However, it is now well-known that gradient descent is not always a good optimizer for NNs, which can partially explain the unsatisfactory practical performance of the NTK regression estimator. In this paper, we introduce the Weighted Neural Tangent Kernel (WNTK), a generalized and improved tool, which can capture an over-parameterized NN's training dynamics under different optimizers.
Theoretically, in the infinite-width limit, we prove: i) the stability of the WNTK at initialization and during training, and ii) the equivalence between the WNTK regression estimator and the corresponding NN estimator with different learning rates on different parameters. With the proposed weight update algorithm, both empirical and analytical WNTKs outperform the corresponding NTKs in numerical experiments.
\end{abstract}

\section{Introduction}
\label{Introduction}

Neural Networks (NNs) are increasingly gaining popularity due to their powerful flexibility in model fitting and their impressive performance in many applications. However, owing to their nested and nonlinear architectures, the theoretical study on NNs is relatively lacking. 
In recent years, the Neural Tangent Kernel (NTK) has been theoretically proved to be an excellent tool to describe a NN in the infinite-width limit using the over-parameterization property \cite{jacot2018neural}. An over-parameterized NN trained by gradient descent is equivalent to the corresponding NTK regression estimator \cite{Arora2019OnEC}. Some regularization methods of NNs can also be transferred into regularization terms of the NTK regression estimator \cite{Hu2019UnderstandingGO}.

Although the NTK has built up a promising bridge between theoretical analysis and practice learning algorithms, the performance gap is still obvious, as shown by Arora et al. \yrcite{Arora2019OnEC}: Accuracy of CNTKs (NTKs that correspond to Convolutional Neural Networks) is always about 5\% lower than that of the corresponding Convolutional Neural Networks (CNNs). Many modifications have been proposed to fulfill this gap \cite{hanin2019finite,chen2020label}. Nevertheless, those modifications generally depart from the initial idea that the NTK is a tool to simulate the NN's training phase dynamics. 


In this paper, we generalize the original NTK to improve its classification performance while keeping a close link to NN training. The original NTK describes the dynamics of the training phase of an over-parameterized NN using gradient descent as the optimizer. However, for an under-parameterized NN, it has been well-known that gradient descent is not always a good optimizer. 
Instead, there have been many efficient algorithms, e.g., SGD with Momentum \cite{sutskever2013importance, cutkosky2020momentum}, AdaGrad \cite{zeiler2012adadelta}, Adam \cite{zhang2018improved}. Those modifications multiply certain adjustment matrices on the gradient descent direction, which should be correspondingly reflected in the NTK. 


Our idea is to introduce weight terms on the NTK, resulting in the Weighted Neural Tangent Kernel (WNTK). Consider a fully connected network with input dimension $d_0 \in {\mathbb{N}^{+}}$ and width $d_l\in {\mathbb{N}^{+}}$ at the $l$-th layer. Let $f^{(l)}$ denote the output of the $l$-th layer of the NN and let all activation functions of different layers be a coordinate-wise one $\sigma$. $P$ is the number of parameters and $\theta \in {\mathbb{R}^P}$ represents all the parameters. For each input sample $x \in \mathbb{R}^{d_0}$, the recurrence relation for this NN architecture is defined as
\begin{equation} 
\label{nnarchitecture}
f^{(l)}(\theta, x) = W^{(l)}\sqrt{\frac{1}{d_{l-1}}} \sigma(f^{(l-1)}(\theta, x)),
\end{equation}
where $W^{(l)} \in {\mathbb{R}}^{d_l \times d_{l-1}}$ is the weight matrix between the $(l-1)$-th and the $l$-th layer. We formally give the WNTK:
\begin{equation} 
\mathcal{A}(x_i, x_j) = \sum_{p=1}^P a_p (\partial_{\theta_{p}} f^{(L)}(\theta, x_i))^{\top} \partial_{\theta_{p}} f^{(L)}(\theta, x_j) , \label{wntkdefinition1}
\end{equation}
where 
$a_p \in {\mathbb{R}}$. Intuitively, the introduced weights could model the adjustment on gradient descent and thus improve the classification performance. Simultaneously, the WNTK reserves two theoretical properties:
i) stability at initialization and during training; ii) equivalence between a WNTK regression estimator and the corresponding NN estimator. These two properties are crucial for linking NNs and NTKs. However, unlike NTKs locked to fixed descent direction, WNTKs can describe different optimizers. 

There are other ways to understand the weight terms.

1. Consider a NN under gradient descent. Theoretical works have shown that techniques, like Dropout and Freezing Layer, can accelerate its convergence, enhance its generalization ability and improve its performance \cite{srivastava2014dropout,brock2017freezeout}. The essence of such techniques is to set different learning rates on different parameters, which is equivalent to applying adjusted gradient descent. 
Therefore, weight terms in the WNTK can also be interpreted as learning rates on parameters. 
A theoretical explanation on this link between weight terms and learning rates could be found in Section \ref{theory}.

2. From the view of kernel learning, the NTK could be regarded as an equal-weighted sum of multiple kernels, each of which describes the gradient on each layers, i.e., 
\begin{equation}
\label{layerwisekernel}
\begin{aligned}
& \  \ \Theta^{(L)}_{l}(x_i,x_j) \\
= & \sum_{\theta_p \ \mathrm{in \ layer} \ l } (\partial_{\theta_{p}} f^{(L)}(\theta, x_i))^{\top}\partial_{\theta_{p}} f^{(L)}(\theta, x_j)
\end{aligned}
\end{equation}
where $x_i, x_j$ are two input samples in the training set. Many discussions on multiple kernel learning \cite{sonnenburg2006large, gonen2011multiple} imply that giving different weights, just like the WNTK
\begin{equation}
\label{layerwisewntk}
\mathcal{A}(x_i,x_j) = \sum_{l=1}^{L}a^L_l\Theta^{(L)}_{l}(x_i,x_j),
\end{equation}
could improve the performance. For these layer-wise weights, we establish an update algorithm to find a good WNTK, see Algorithm \ref{alg:algo1}. One can also element-wisely adjust the weights, which falls into the idea of non-parametric kernel learning \cite{zhuang2011family,liu2020learning} that learns the element values. 

To summarize, this paper proposes the WNTK, analyzes its properties, designs a learning algorithm, and evaluates the performance of both empirical and analytic kernels. The main contributions are listed below:

1. We introduce the Weighted Neural Tangent Kernel, a generalized and improved tool to simulate over-parameterized NN dynamics. Compared to the NTK, the WNTK could describe the dynamics of the NN trained by different optimizers. 

2. In the infinite-width limit, we prove that the WNTK is stable both at initialization (\textbf{Theorem \ref{t1}}) and during the corresponding NN training phase (\textbf{Theorem \ref{t2}}) under adjusted gradient descent.

3. In the infinite-width limit, we prove the equivalence between a WNTK regression estimator and the corresponding NN estimator under adjusted gradient descent (\textbf{Theorem \ref{t4}}).

4. We design a weight update algorithm (\textbf{Algorithm} \ref{alg:algo1}) and verify in the experiment that the WNTK does have significant improvement from the NTK.

The rest of the paper is organized as follows.
Section \ref{Related Work} summarizes the research background and the related work.
Section \ref{preliminary} reviews the derivation of the NTK and introduces the notations. 
Section \ref{theory} reveals the connection between the WNTK and NN training dynamics.
Section \ref{algorithm} establishes a 
weight update algorithm for learning a good WNTK.
Section \ref{experiment} presents experimental results. 

\section{Related Work}
\label{Related Work}

Initially, the correspondence between wide NNs and kernel machines was noted by Neal's pioneering work \yrcite{neal2012bayesian}, which was followed by a series of papers \cite{williams1997computing,lee2017deep,hazan2015steps,matthews2018gaussian,novak2018bayesian,garriga2018deep}. Meanwhile, over-parameterized NNs showed stunning performance and strong generalization ability in practice, which gave rise to many discussions on the nature of over-parameterization \cite{arora2019fine,allen2018learning,chizat2018lazy,li2018learning,ghorbani2019limitations,arora2018optimization}. It has been shown that, under the infinite width assumption, the evolution of a wide NN of any depth possesses the same dynamics as a linearized model \cite{lee2019wide}. Based on all these discussions, several different NN-induced kernels were proposed as tools to explain the behavior of NNs \cite{daniely2016toward,shankar2020neural,anselmi2015deep}, where belong the NTK \cite{jacot2018neural} and its derivations \cite{Arora2019OnEC,li2019enhanced,du2019graph,alemohammad2020recurrent}. The most remarkable property of the NTK is that the NTK predictor is equivalent to the corresponding NN predictor trained by gradient descent. As the inner connection between kernel machines and NNs is progressively being revealed \cite{daniely2016toward,belkin2018understand}, the NTK is considered as a bridge between NNs and traditional kernel machines.

Although Du et al. \yrcite{du2018gradient} prove that gradient descent optimizes over-parameterized wide NNs, the performance of the NTK in practice is relatively unsatisfactory compared to NNs \cite{samarin2020empirical,ghorbani2019linearized,wei2019regularization}. The NTK can outperform the corresponding NN only on small datasets \cite{arora2019harnessing}. On large datasets, there exists a noticeable performance gap between them \cite{Arora2019OnEC}, which triggers many works on improving the NTK's performance, like adding label-awareness \cite{chen2020label}, conducting finite-width correction \cite{hanin2019finite}, and creating neural tangent hierarchy \cite{huang2020dynamics}.  

The main idea of this paper is to generalize the NTK to discuss different optimizers rather than gradient descent. It has been intensively discussed that Adam \cite{zhang2018improved}, AdaGrad \cite{zeiler2012adadelta}, SGD with Momentum \cite{sutskever2013importance,cutkosky2020momentum}, Etc, have better performance than gradient descent, at least on under-parameterized NNs.



\section{Recap of NTK}
\label{preliminary}

Let us focus on the NN architecture described by (\ref{nnarchitecture}). Consider an $L$-layer NN. The corresponding NTK is defined as follows: 
\begin{equation}
\label{ntk}
\begin{aligned}
\Theta(x_i, x_j) &= \Theta^{(L)}(\theta) (x_i, x_j) \\
&= \sum_{p=1}^P ( \partial_{\theta_{p}} f^{(L)}(\theta, x_i))^{\top}\partial_{\theta_{p}} f^{(L)}(\theta, x_j) ,
\end{aligned}
\end{equation}
The two most significant works on the NTK are: i) It converges in probability to a closed-form kernel at initialization; ii) It stays constant during training if the NN is sufficiently wide \cite{jacot2018neural}. 
By default, $\Theta$ and $\mathcal{A}$ refer to the NTK and the WNTK on the training set, and $ \dot\sigma $ represents the derivative of $ \sigma$. The following formula shows the stable closed-form kernel induced by an $L$-layer dense network
\begin{equation}
\label{ntkstability1}
\begin{aligned}
\Theta \to \Theta^{(L)}_{\infty} \otimes Id_{d_L},
\end{aligned}
\end{equation}
where $\otimes$ stands for the Kronecker product, $Id_{d_L} $ denotes the identity matrix of size $d_L$, and $\Theta^{(L)}_{\infty}$ is defined recursively by
\begin{align*}
\Theta^{(1)}_{\infty}(x_i, x_j) =& \ \Sigma^{(1)}(x_i, x_j), \\
\Theta^{(l)}_{\infty}(x_i, x_j) =& \ \Theta^{(l-1)}_{\infty}(x_i, x_j)\dot \Sigma^{(l)}(x_i, x_j)\\ 
& \ + \Sigma^{(l)}(x_i, x_j),\\
\Sigma^{(1)}(x_i, x_j) =& \ \frac{1}{d_0}x^{\top}_ix_j,\\
\Sigma^{(l)}(x_i, x_j) =& \ \mathbb{E}_{f \sim \mathcal{N}(0, \Sigma^{(l-1)})}[\sigma(f(x_i))\sigma(f(x_j))],\\
\dot \Sigma^{(l)}(x_i, x_j) =& \ \mathbb{E}_{f \sim \mathcal{N}(0, \Sigma^{(l-1)})}[\dot\sigma(f(x_i))\dot\sigma(f(x_j))].
\end{align*}
Under the assumption that $d_1 =...=d_{L-1}=d$ and $d_L=1$, the following formula reflects the stability of the NTK during the training phase of an over-parameterized NN, or equivalently, the convergence between the NTKs $\Theta_{0}$ and $\Theta_{t}$ at moment $0$ and $t$:

\begin{equation}
\label{ntkstability2}
\begin{aligned}
\forall t, ||\Theta_{0} - \Theta_{t}||_{F} \to 0, as \ d \to \infty.
\end{aligned}
\end{equation}

Under the infinite width assumption, the fully trained NN estimator $f^{\star}$ can be proved to be equivalent to a NTK regression estimator \cite{jacot2018neural}, i.e., 
\begin{equation}
\label{ntkequivalence}
\begin{aligned}
f^{\star}(\theta, x) \to \Theta(x, \mathcal{X})^{\top}(\Theta(\mathcal{X}, \mathcal{X}))^{-1}\mathcal{Y},
\end{aligned}
\end{equation}
where $(\mathcal{X},\mathcal{Y})$ stands for the training set of size $n$ and $x$ is a sample.

\section{Theory: WNTK and Neural Network Training dynamics}
\label{theory}
Intuitively, the WNTK is a generalization of the NTK and in this section we will show that this generalization enables us to describe the training process under adjusted gradient descent using different optimizers. 
For the NN architecture described in (\ref{nnarchitecture}), we reshape the formula of the WNTK (\ref{wntkdefinition1}) as the following,
\begin{equation}
\label{wntkdefinition2}
\begin{aligned} 
\mathcal{A}(x_i, x_j) = (\partial_{\theta} f^{(L)}(\theta, x_i))^{\top} (a \odot \partial_{\theta} f^{(L)}(\theta, x_j)).
\end{aligned}
\end{equation}
If we assume $d_L=1$, we have,
\begin{equation}
\label{wntkdefinition3}
\begin{aligned} 
\mathcal{A} = \mathcal{A}(\mathcal{X},\mathcal{X}) = \nabla_{\theta} f(\mathcal{X}) (a \odot \nabla_{\theta} f(\mathcal{X})^\top),
\end{aligned}
\end{equation}
where $a = [a_1, a_2, ... ,a_P]^\top$, $a_p \in {\mathbb{R}}$,  $\nabla_{\theta} f(\mathcal{X}) \in {\mathbb{R}}^{n \times P}$ represents the gradient of $f$ on $\theta$ with respect to the training set $\mathcal{X}$, and $\odot$ refers to the penetrating face product.

\subsection{Stability Analysis of WNTK}

In this subsection, we theoretically show that the WNTK captures the behavior of an over-parameterized NN fully-trained by a non-equally weighted optimizer. In the infinite-width limit, the WNTK, of which the corresponding NN is randomly initialized, converges in probability to a deterministic limit.

\begin{theorem}
\label{t1}
    \textbf{(Stability at Initialization).} \textit{Consider a network of depth L at initialization, with a Lipschitz nonlinearity $\sigma$. Suppose that, for every parameter $\theta_p$ in the $l$-th layer, its weight term in the WNTK follows a layer-wise distribution $\rho_l$. In the limit as the layer width $d_1,...,d_{L-1}$ tends to infinity sequentially, the layer-wise WNTK converges in probability to a deterministic limiting kernel}
    \begin{equation}
    \label{wntkstability1}
    \begin{aligned}
    \mathcal{A} \to \mathcal{A}^{(L)}_{\infty} \otimes Id_{d_L}.
    \end{aligned}
    \end{equation}
    \textit{With $\mu_l$ referring to the expectation of all element-wise weights which follow the layer-wise distribution $\rho_l$, the scalar kernel $\mathcal{A}^{(L)}_{\infty}: {\mathbb{R}}^{d_0} \times {\mathbb{R}}^{d_0} \to \mathbb{R}$ is defined recursively by}
    \begin{align*}
    \mathcal{A}^{(1)}_{\infty}(x_i, x_j) =& \ \mu_1\Sigma^{(1)}(x_i, x_j), \\
    \mathcal{A}^{(l)}_{\infty}(x_i, x_j) =& \ \mathcal{A}^{(l-1)}_{\infty}(x_i, x_j) \dot \Sigma^{(l)}(x_i, x_j)\\
    & + \mu_{l}\Sigma^{(l)}(x_i, x_j),
    \end{align*}
    \textit{where}
    \begin{align*}
    \Sigma^{(1)}(x_i, x_j) &= \frac{1}{d_0}x^{\top}_ix_j,\\
    \Sigma^{(l)}(x_i, x_j) &= \mathbb{E}_{f \sim \mathcal{N}(0, \Sigma^{(l-1)})}[\sigma(f(x_i))\sigma(f(x_j))],\\
    \dot \Sigma^{(l)}(x_i, x_j) &= \mathbb{E}_{f \sim \mathcal{N}(0, \Sigma^{(l-1)})}[\dot\sigma(f(x_i))\dot\sigma(f(x_j))].
    \end{align*}
\end{theorem}

\textbf{Proof Sketch.} The demonstration is inspired by Jacot et al. \yrcite{jacot2018neural} and is by induction. Take  $\mathcal{A}^{(l)}$ the WNTK of an $l$-layer fully-connected NN. For the first hidden layer, it can be easily demonstrated that $\mathcal{A}^{(1)} \to \mathcal{A}^{(1)}_{\infty} \otimes Id_{d_1} $. For layer $l>1$, we divide $\mathcal{A}^{(l)}$ into two parts: the sum of derivatives on parameters in the first $(l-1)$ hidden layers and the sum of derivatives on parameters in the $l$-th layer. In the first case, as $\mathcal{A}^{(l-1)} \to \mathcal{A}^{(l-1)}_{\infty} \otimes Id_{d_{l-1}}$ is assumed to be correct by induction hypothesis, $\mathcal{A}^{(l)}_{\mathrm{first} \ (l-1) \ \mathrm{layers}} \to \mathcal{A}^{(l-1)}_{\infty}(x_i, x_j) \dot \Sigma^{(l)}(x_i, x_j)$ can be derived. In the second case, it can be easily demonstrated that $\mathcal{A}^{(l)}_{l\mathrm{-th} \ \mathrm{layer}} \to  \mu_l\Sigma^{(l)}(x_i, x_j)$. Thus, we have proved the whole theorem. A detailed proof can be found in Appendix.

Note that (\ref{ntkstability1}) and (\ref{wntkstability1}) represent one of the key properties possessed by the NTK and the WNTK, respectively. Similar to $\Theta^{(L)}_{\infty}$,  $\mathcal{A}^{(L)}_{\infty}$ only depends on the choice of activation function, the depth of the NN, and the variance of the parameters at initialization.

Next, we study the stability of the WNTK of a standardly parameterized NN when applying full-batch gradient descent under Assumptions 1-4 (see below). The demonstration is built on the recent work by Lee et al. \yrcite{lee2019wide}, which proves the stability of the NTK during the training phase of an over-parameterized NN. 

\textbf{Standard Parameterization:} A standard parameterized NN is formulated as
\begin{align*}
\begin{cases}
h^{l+1} = x^{l}W^{l+1}\\
x^{l+1} = \sigma(h^{l+1})
\end{cases}
\end{align*}
with $x^{l}$ being the output of the $l$-th layer, $x^{l+1}$ the output of the $(l+1)$-th layer, and each parameter $w^{l}_{i,j}$ in the connection matrices $W^{l}$ initialized from i.i.d. Gaussians $\mathcal{N}(0, \frac{\sigma^{2}_w}{d})$. In the following, we assume $d_L = 1$.

\textbf{Assumptions 1-4:} 
\begin{itemize}
\item The widths of the hidden layers are identical, i.e. $d_1 = ... = d_{L-1} = d$.
\item The analytic WNTK is full-rank. And we set $\eta_{\mathrm{critical}}$ $= 2(\lambda_{\min}+\lambda_{\max})^{-1} > 0 $. $\lambda_{\min}$ and $\lambda_{\max}$ are the smallest and the largest eigenvalues of the WNTK. $\lambda_{\max} \ge \lambda_{\min} > 0$.
\item The training set $(\mathcal{X},\mathcal{Y})$ is contained in some compact sets and $x_i \ne x_j$ for all $x_i,x_j \in \mathcal{X}$.
\item The activation function $\sigma$ satisfies: $|\sigma(0)|$, $||\dot\sigma||_{\infty}$, $\sup_{x_i \ne x_j}|\dot\sigma(x_i) - \dot\sigma(x_j)|/|x_i-x_j| < \infty $.
\end{itemize}

Assumption 2 is satisfied when all inputs are limited to the unit sphere. See the proof of the positive-definiteness of the NTK in Jacot et al. \yrcite{jacot2018neural}. A similar conclusion can be derived for the WNTK.

\begin{theorem}
\label{t2}
    \textbf{(Stability during Training).} \textit{Under Assumptions 1-4 and suppose that an  L-layer neural network $f$ with random initialization is trained by gradient decent with different learning rate $\eta a_p$ on different parameter $\theta_p$,  where $\eta=\frac{\eta_0}{d}$. 
    For $\delta_0 >0$ and $\eta_0<\eta_{\mathrm{critical}}$, there exist $R_0 > 0$, $N \in \mathbb{N}$, and $K>1$ such that for every $d \ge N$, 
    we have the following inequality with probability at least $(1-\delta_0)$:}
    \begin{equation}
    \label{wntkstability2}
    \begin{aligned}
    ||\mathcal{A}_{0} - \mathcal{A}_{t}||_{F} \leq \frac{6K^{3}R_0a_{\max}^2}{\lambda_{\min}}d^{-\frac{1}{2}}, \forall t,
    \end{aligned}
    \end{equation}
    \textit{where $\mathcal{A}_0$ and $\mathcal{A}_t$ are the WNTK at moment $0$ and $t$ under standard parameterization, and $a_{\max} = \max\{a_i| i = 1,...,P\}$.}
\end{theorem}

\textbf{Proof Sketch.} We begin with the definition of $\mathcal{A}$ (\ref{wntkdefinition3}). By applying local lipschitzness properties of the Jacobean (\textbf{Lemma} 1 in Appendix), we find that a sufficient condition of $||\mathcal{A}_{0} - \mathcal{A}_{t}||_{F} \leq \frac{6K^{3}R_0a_{\max}^2}{\lambda_{\min}}d^{-\frac{1}{2}}$ is $||\theta_0-\theta_t||_2 \leq \frac{3KR_0a_{\max}}{\lambda_{\min}}d^{-\frac{1}{2}}$. Secondly, we deduce by induction that the latter inequality can be attained if $||f(\theta_t, \mathcal{X}) - \mathcal{Y}||_2\leq (1-\frac{\eta_0\lambda_{\min}}{3})^\top R_0$. Thirdly, we prove the correctness of the last inequality under Assumptions 1-4. A detailed proof can be found in Appendix.

For the WNTK, (\ref{wntkstability2}) shows the stability of the WNTK during training, playing similarly the rule of (\ref{ntkstability2}) in the NTK analysis. 
Adjustment on descent direction does not affect the stability of the WNTK. So, we find no dependence between the stability of the WNTK and the optimizer of an over-parameterized NN. The weight terms in the WNTK can characterize different optimizers.

\subsection{Equivalence between WNTK Regression Estimator and NN Estimator using Different Optimizer}
In this subsection, we will show that a standard parameterized wide NN estimator under adjusted gradient descent is equivalent to its related WNTK regression estimator. Here we consider sum of squares as the loss function, i.e., $ (x_i,y_i) \in (\mathcal{X},\mathcal{Y})$,
\begin{align*} 
\mathcal{L}(f(\theta, \mathcal{X})) = \frac{1}{2}\sum_{i=1}^{n}(f(\theta,x_i) - y_i)^2.
\end{align*}
First, under adjusted gradient descent, an over-parameterized NN still drops into the lazy regime, and thus can be approximated by a linearized network defined as follows,
\begin{equation}
    \label{linearizedmodel}
    \begin{aligned}
    f^{\mathrm{lin}}_t(x) = f(\theta_0,x) + \nabla_{\theta} f(x)|_{\theta = \theta_0}(\theta_{t}-\theta_0).
    \end{aligned}
\end{equation} 
$f^{\mathrm{lin}}_t(x)$ and $\theta_{t}$ denote the value of $f^{\mathrm{lin}}(x)$ and $\theta$ at moment $t$, respectively. The dynamics of adjusted gradient flow using this linearized model are governed by
\begin{equation}
\label{dynamic1}
\dot \theta_t = - \eta (a \odot \nabla_{\theta} f_0(\mathcal{X})^{\top}) \nabla_{f^{\mathrm{lin}}_t(\mathcal{X})}\mathcal{L},
\end{equation} 
\begin{equation}
\label{dynamic2}
\dot f^{\mathrm{lin}}_t(x) = - \eta \mathcal{A}_0(x,\mathcal{X})\nabla_{f^{\mathrm{lin}}_t(\mathcal{X})}\mathcal{L}.
\end{equation} 

\begin{theorem}
\label{t3}
    \textbf{(WNTK Lazy Regime).} \textit{Under Assumptions 1-4, applying gradient descent with different learning rate $\eta a_p$ on different parameter $\theta_p$, for every $x \in  {\mathbb{R}}^{d_0}$ satisfying $||x||_2 \leq 1$, for all $t\ge0$, as $ d \to \infty$, with probability arbitrarily close to 1 over random initialization, we have}
    \begin{equation}
    \label{wntklinear}
    \begin{aligned} 
    || f(\theta_t, \mathcal{X}) - f^{\mathrm{lin}}_t(\mathcal{X})||_{2} \to 0,\\
    || f(\theta_t, x) - f^{\mathrm{lin}}_t(x)||_{2} \to 0.
    \end{aligned}
    \end{equation}
\end{theorem}

The proof of this theorem follows a similar proof in Lee et al. \yrcite{lee2019wide}. The details could be found in Appendix.

\begin{theorem}
\label{t4}
    \textbf{(Equivalence between WNTK Regression and Neural Network).} \textit{Fix a set of learning rate $ \eta a > 0 $. Consider gradient descent of an infinitely wide neural network with initialization $\theta(0)$. If $ \eta $ is bounded by a known value $ \eta_{\mathrm{c}}$ and the neural network is defined to have a very small initial output, the fully trained NN estimator $f^{\star}$ is equivalent to the WNTK ridge regression estimator with regularization term going to zero:}
    \begin{equation}
    \label{wntkequivalence}
    \begin{aligned} 
    f^{\star}(\theta, x) = \mathcal{A}(x, \mathcal{X})^\top(\mathcal{A}(\mathcal{X},\mathcal{X}))^{-1}\mathcal{Y},
    \end{aligned}
    \end{equation}
    where $\mathcal{A}(\mathcal{X}, \mathcal{X}) \in \mathbb{R}^{n \times n}$ stand for the WNTK on the training set and $\mathcal{A}(x, \mathcal{X}) = [\mathcal{A}(x, x_1), ... , \mathcal{A}(x, x_n)]^\top \in \mathbb{R}^{n}$.
\end{theorem}

\textbf{Proof.} Let us first review this NN's adjusted gradient descent dynamics. At iteration $k$, adjusted gradient descent on $\theta$ with learning rate $a$ iterates as follows:
\begin{equation}
\label{wntkiteration}
\begin{aligned} 
\theta(k+1) = \theta(k) - \eta( a\odot \nabla_{\theta}\mathcal{L}(\theta(k))), a \in {\mathbb{R}}^{P}.
\end{aligned}
\end{equation}
Under the infinite-width assumption, according to the property of lazy regime (\ref{wntklinear}), as $d \to \infty$:
\begin{equation}
\label{t4.1}
\begin{aligned} 
\mathcal{L}(f(\theta, \mathcal{X})) = & \  \frac{1}{2}\sum_{i=1}^{n}(f(\theta(0),x_i)+\\
& \ \left<\nabla_{\theta} f(\theta(0),x_i), \theta - \theta(0) \right> - y_i)^2.
\end{aligned}
\end{equation}
We substitute (\ref{t4.1}) into (\ref{wntkiteration}) and we get
\begin{align*} 
& \ \theta(k+1)\\
=& \ \theta(k) - \eta( a\odot(\sum_{i=1}^{n}(f(\theta(k),x_i) - y_i)\nabla_{\theta}f(\theta(0),x_i)))\\
=& \ \theta(k) - \eta(a\odot(\sum_{i=1}^{n}(f(\theta(0),x_i)+\\
& \ \left<\nabla_{\theta} f(\theta(0),x_i), \theta(k) - \theta(0) \right> - y_i)(\nabla_{\theta}f(\theta(0),x_i))))\\
=& \ \theta(k) - \eta (a\odot U)L - \eta (a\odot U)U^{T}(\theta(k) - \theta(0)),
\end{align*}
where
\begin{align*} 
& L = \ (l_0(0),l_1(0),...,l_n(0))^T \in {\mathbb{R}}^{n},\\
& l_i(0)= \ f(\theta(0),x_i) - y_i \in {\mathbb{R}},\\
& U = \ (\nabla_{\theta} f_t(\mathcal{X})|_{\theta = \theta_0})^\top \in {\mathbb{R}}^{P \times n}.
\end{align*} 
Let $c(k)$ $= \theta(k)$ $+ (\eta (a\odot U)U^T)^{-1}$ $(\eta (a\odot U)L - \eta  (a\odot U)U^T\theta(0)) $. We find the following iterative relation
\begin{align*} 
c(k+1) = (Id - \eta (a\odot U)U^T)c(k).
\end{align*} 
Set $ \eta_{\mathrm{c}} = \frac{1}{|(a\odot U)U^T|}$. According to (\ref{wntkdefinition3}), we can deduce, as $k\to \infty$,
\begin{align*} 
\theta(k) =& \ \theta(0) -(Id - (Id - \eta (a\odot U)U^T)^k)\\
& \ (\eta (a\odot U)U^T)^{-1}\eta (a\odot U)L\\
=& \ \theta(0) - (\eta (a\odot U)U^T)^{-1}\eta (a\odot U)L\\
=& \ \theta(0) - (a\odot U)(\mathcal{A})^{-1}L.
\end{align*} 
As we assume that the neural network is defined to have a very small initial output, the predictor of this wide neural network is:
\begin{align*} 
f^{\star}(\theta, x) =& \ f^{lin}_{\infty}(x) = \left<\nabla_{\theta} f(\theta(0),x), \theta - \theta(0) \right>\\
=& \ -(\nabla_{\theta} f_t(x)|_{\theta = \theta_0})^{\top}(a\odot U)(\mathcal{A})^{-1}L  \\
=& \ \mathcal{A}(x, \mathcal{X})^\top(\mathcal{A})^{-1}\mathcal{Y}. 
\end{align*} 
$\hfill\blacksquare$

Note that for an empirical WNTK of a finite-width NN, we can no longer ignore the initial output and the prediction takes the following form:
\begin{equation}
\label{finitewidth}
    \begin{aligned}
    f^{\star}(\theta, x) \simeq & \ \mathcal{A}(x, \mathcal{X})^\top(\mathcal{A})^{-1}\mathcal{Y} + (f(\theta_0, x)  \\
    & \ - \mathcal{A}(x, \mathcal{X})^\top(\mathcal{A})^{-1}f(\theta_0, \mathcal{X})).
    \end{aligned}
\end{equation}
Moreover, regularization in the WNTK ridge regression is equivalent to regularization in over-parameterized NNs. Such properties have already been discussed on the NTK \cite{wei2019regularization}.

Below is an example, where WNTK simulates
the training process of a NN under gradient descent with momentum. Take $\beta \in [0,1]$. At iteration $k$, the gradient descent with momentum is given by the following,
\begin{align*} 
\Delta \theta =& -\eta \nabla_{\theta} f(\theta(k), \mathcal{X})\\
&- \eta \beta (\nabla_{\theta} f(\theta(k), \mathcal{X}) -\nabla_{\theta} f(\theta(k-1), \mathcal{X})).
\end{align*} 
Suppose that this NN has only one output at the final layer. Choose the following weights in WNTK,
\begin{align*} 
a_p = 1 + \beta (\nabla_{\theta_p}f(\theta(0), \mathcal{X}))^{-1} H_p(\theta(0)), 
\end{align*} 
where $	H_p(\theta(0), \mathcal{X})$ refers to the Hessian Matrix at $\theta(0)$ with respect to $\theta_p$, namely, the derivative gradient information at the initial point $\theta(0)$. Analytically, it can be approximated by the second derivative of $f$. Practically, it can be obtained by training the neural network for one epoch using full-batch gradient descent. In this case, the WNTK represents the same dynamics as gradient descent with momentum:
\begin{align*} 
\Delta \theta_p &= - \eta(a \odot \nabla_{\theta_p} f(\theta(0), \mathcal{X})) \\
&= - \eta(\nabla_{\theta_p} f(\theta(0), \mathcal{X}) + \beta H_p(\theta(0), \mathcal{X})).
\end{align*} 
Gradient descent with momentum and its derives has been revealed to be useful and effective on over-parameterized NN \cite{liu2018accelerating}. The above example shows that such improvements on optimizer can be equivalently transferred to WNTK regression estimators. 

\section{Algorithm: Weight Update}
\label{algorithm}

In the previous sections, we show the equivalence between the WNTK regression estimator and the corresponding over-parameterized NN estimator under the adjusted descent direction.
Because of the outperformance of other optimizers over gradient descent, the WNTK is promising to have better performance than the NTK if we can find a good descent direction, i.e., good weights. One problem here is that WNTK, also NTK, are certainly overfitting on the training set. To achieve better generalization performance, we have tried cross-validation, kernel alignment, regularization rules, Etc, and find that the following update rule is the most suitable. 



\textbf{Gradient Descent on Layer-wise Weight.} One way to find the optimal layer-wise weights is to perform gradient descent on weights. The original training data can be divided into a training set $(\mathcal{X}_{\mathcal{T}}, \mathcal{Y}_{\mathcal{T}})$ and a validation set $(\mathcal{X}_{\mathcal{V}}, \mathcal{Y}_{\mathcal{V}})$. We use $\mathcal{X}_{\mathcal{T}}$ to generate the WNTK estimator and use its prediction on $\mathcal{X}_{\mathcal{V}}$ to update its layer-wise weights. The weights could be updated layer-wisely or element-wisely. 
Consider an $L$-layer NN and $d_L=1$. With the definition of layer-wise WNTK (\ref{layerwisekernel}) and (\ref{layerwisewntk}), the layer-wise gradient of a weight term $a^{L}_i$ at the $l$-th layer is given as follows,
\begin{align*}
& \ \frac{\partial f^{\star}}{\partial a^{L}_i}(\theta, \mathcal{X}_{\mathcal{V}}) \\
= \ & \frac{\partial\mathcal{A}_{\mathcal{V}}^\top}{\partial a^{L}_i}\mathcal{A}^{-1}\mathcal{Y}_{\mathcal{T}} + \mathcal{A}_{\mathcal{V}}^\top\frac{\partial \mathcal{A}^{-1}}{\partial a^{L}_i}\mathcal{Y}_{\mathcal{T}}\\
= \ & ({\Theta_{\mathcal{V}}}^{(L)}_{l})^\top\mathcal{A}^{-1}\mathcal{Y}_{\mathcal{T}} - \mathcal{A}_{\mathcal{V}}^\top\mathcal{A}^{-1}\Theta^{(L)}_{l}\mathcal{A}^{-1}\mathcal{Y}_{\mathcal{T}},
\end{align*}
where 
\begin{align*}
{\Theta_{\mathcal{V}}}^{(L)}_{l} &= \sum_{\theta_p \ \mathrm{in \ layer} \ l } \partial_{\theta_{p}} f^{(L)}(\theta, \mathcal{X}_{\mathcal{T}})^{\top} \partial_{\theta_{p}} f^{(L)}(\theta, \mathcal{X}_{\mathcal{V}})
\end{align*}
and $\mathcal{A}_\mathcal{V}$ represents the WNTK between $\mathcal{X}_{\mathcal{T}}$ and $\mathcal{X}_{\mathcal{V}}$.
The whole algorithm could be found in Algorithm~\ref{alg:algo1}. Notice that 
the above closed-form formula enables us to avoid calculating the gradient through the Auto-Gradient method in packages like Tensorflow or Pytorch, which is of high cost. Instead, by storing the layer-wise kernels $\Theta^{(L)}_{l}$ and $\mathcal{A}^{-1}$ in memory, the calculation of gradients of layer-wise weights becomes much lighter. The time cost of this algorithm is mainly from the matrix inversion operation $\mathcal{A}^{-1}$, which is of $\mathcal{O}(n^3)$. So the time complexity of each gradient update iteration is $\mathcal{O}(Ln^3)$. Its space complexity is $\mathcal{O}(Ln^2)$, which mainly comes from the storage of $\mathcal{A}^{-1}$ and $\Theta^{(L)}_{l}$, $l = 1,..,L$.

With little modification, the above update is applicable to train element-wise weights. 
In our experiment, we choose layer-wise update strategy, since it can better control complexity. Moreover, with different layer-wise weights, the WNTK could explain how layer-wise information is integrated and why one needs different learning rates for training NNs, which has been reported in many practical papers \cite{takase2018effective,liu2019variance}.


To investigate the meaning of layer-wise weights, we consider a fully connected NN at infinite width and focus on two successive layers. Reformulate the layer-wise kernel (\ref{layerwisekernel}) as the following,
\begin{equation}
\label{layerwisekernel2}
\Theta^{(L)}_{l}(x_i,x_j) = \Sigma^{(l)}(x_i,x_j)\prod_{l'=l+1}^{L}\dot\Sigma^{(l')}(x_i,x_j).
\end{equation}
The activation function is supposed to be $ReLU$ and we have
\begin{align*}
& \ \mathbb{E}_{(u,v) \sim \mathcal{N}(0, D\Lambda^{(l)} D)}[\sigma(u)\sigma(v)] \\
=& \ \frac{\lambda(\pi-\arccos(\lambda))+\sqrt{1-\lambda^2}}{2\pi}c_1c_2,\\
& \ \mathbb{E}_{(u,v) \sim \mathcal{N}(0, D\Lambda^{(l)} D)}[\dot\sigma(u)\dot\sigma(v)] \\
=& \ \frac{\pi-\arccos(\lambda)}{2\pi},
\end{align*}
where $|\lambda|<1$,
\begin{align*}
D = \begin{bmatrix}
c_1 & 0 \\
0 & c_2
\end{bmatrix} \mathrm{, \ and \ }
\Lambda^{(l)} = \begin{bmatrix}
1 & \lambda \\
\lambda & 1
\end{bmatrix}
\end{align*} 
under the following constraint,
\begin{align*}
D\Lambda^{(l)} D = \begin{bmatrix}
\Sigma^{(l-1)}(x_i, x_i) & \Sigma^{(l-1)}(x_i,x_j) \\
\Sigma^{(l-1)}(x_j, x_i) & \Sigma^{(l-1)}(x_j, x_j)
\end{bmatrix}.
\end{align*}
Note that the analytic kernel $\mathcal{A}^{(L)}_{\infty}$ in (\ref{wntkstability1}) and the WNTK are well determined, once we fix the activation function, the depth of the NN, and the variance of the parameters at initialization. Take a look at the difference between $\Theta^{(L)}_{l}(x_i,x_j)$ and $\Theta^{(L)}_{l-1}(x_i,x_j)$, which is equal to the difference between two WNTKs of the same NN with different layer-wise weights
$
\begin{cases}
a_{i} = 0, \ i\neq l\\
a_{l} = 1
\end{cases}
$
and 
$
\begin{cases}
a_{i} = 0, \ i\neq l+1\\
a_{l+1} = 1
\end{cases}
$.\\
By applying (\ref{layerwisekernel2}), this difference takes the following form:
\begin{align*}
& \ \Theta^{(L)}_{l-1}(x_i,x_j) - \Theta^{(L)}_{l}(x_i,x_j) \\
=& \ (\Sigma^{(l-1)}(x_i,x_j) \dot\Sigma^{(l)}(x_i,x_j) - \Sigma^{(l)}(x_i,x_j))C\\
=& \ \frac{\sqrt{1-\lambda^2}c_1c_2}{2\pi}C\\
=& \ \sqrt{\Sigma^{(l-1)}(x, x)\Sigma^{(l-1)}(x', x') - \Sigma^{(l-1)}(x_i,x_j)^2}C,
\end{align*}
where $C = \prod_{l'=l+1}^{L}\dot\Sigma^{(l')}(x_i,x_j)$ is a constant and is not related to the $(l-1)$-th either the $l$-th layer's output. It is apparent that, compared to $\Theta^{(L)}_{l}(x_i,x_j)$, $\Theta^{(L)}_{l-1}(x_i,x_j)$ possesses additional information related to the similarity of inputs at the $(l-1)$-th layer. Hence, each layer represents similar but not identical information. So, the layer-wise weight term, by nature, should be non-equally weighted. 
It is reasonable to expect that a network's optimal kernel has different weights on different layer-wise kernels.

\begin{algorithm}[tb]
   \caption{Gradient Descent on Layer-wise Weight}
   \label{alg:algo1}
\begin{algorithmic}
   \STATE {\bfseries Input:} training data $\{(\mathcal{X},\mathcal{Y})\}$, initial layer-wise weight $a^L = [a^L_1, a^L_2, ... ,a^L_L]^\top$, train-validation ratio $r$, learning rate $\eta_w$.
   \REPEAT
   \STATE Split $\{(\mathcal{X},\mathcal{Y})\}$ randomly into to a training set $(\mathcal{X}_{\mathcal{T}}, \mathcal{Y}_{\mathcal{T}})$ and a validation set $(\mathcal{X}_{\mathcal{V}}, \mathcal{Y}_{\mathcal{V}})$ with respect to the train-validation ratio $r$. Generate the WNTK $\mathcal{A}$ and the estimator $f^{\star}$ from $\mathcal{X}_{\mathcal{T}}$ and $a^{L}$.
   \FOR{$i=1$ {\bfseries to} $L$}
   \STATE $\Delta_i = 2  (f^{\star}(\theta,\mathcal{X}_{\mathcal{V}}) - \mathcal{Y}_{\mathcal{V}})^{\top} \frac{\partial f^{\star}}{\partial a^{L}_i}(\theta, \mathcal{X}_{\mathcal{V}})$
   \STATE Update $a^{L}_i = a^{L}_i - \eta_w \times \Delta_i $.
   \ENDFOR
   \UNTIL{Stopping criteria is reached.}
\end{algorithmic}
\end{algorithm}

\section{Experiment}
\label{experiment}

This section provides empirical support that the proposed WNTK, also the Weighted Convolutional Neural Tangent Kernel (WCNTK), does improve the classification performance. We will first evaluate the performance of empirical kernels on CIFAR-10, which is an image task. Then we compare the classification performances between the analytic WNTK and NTK on 10 UCI datasets, which are all non-image.
The empirical kernel matrices and NNs are calculated and trained using the Tensorflow package, with a NVIDIA RTX 2080Ti GPU. All the codes could be found from \url{https://github.com/ASTAugustin/ICML_WNTK}.


\begin{table*}[t]
\caption{Binary-classification accuracy on CIFAR-10 dataset. }
\label{table1}
\vskip 0.15in
\begin{center}
\begin{small}
\begin{sc}
\begin{tabular}{ccccccc}
\toprule
dataset & Size & NTK\_pur & WNTK\_pur & NTK\_ini & WNTK\_ini & NN\\
\midrule
Air/Auto    & 100           & 71.33$\pm$1.68    & 76.45$\pm$1.81            & 77.33$\pm$2.22    & \textbf{79.55$\pm$3.30}   & 78.30$\pm$2.03            \\
            & 1000          & 84.22$\pm$0.91    & 85.34$\pm$0.87            & 85.75$\pm$0.87    & 86.98$\pm$1.01            & \textbf{87.82$\pm$0.55}   \\
            & 10000         & 88.76$\pm$0.77    & 89.66$\pm$0.65            & 92.70$\pm$0.80    & 93.61$\pm0.63$            & \textbf{94.29$\pm$0.54}   \\
Cat/Bird    & 100           & 59.96$\pm$2.98    & 64.37$\pm$3.29            & 62.11$\pm$3.57    & \textbf{65.41$\pm$2.37}   & 56.87$\pm$2.22            \\
            & 1000          & 72.22$\pm$1.62    & \textbf{73.67$\pm$1.06}   & 69.78$\pm$1.74    & 71.14$\pm$1.36            & 67.38$\pm$6.09            \\
            & 10000         & 76.15$\pm$0.35    & 76.67$\pm$0.58            & 78.00$\pm$1.81    & \textbf{79.15$\pm$1.62}   & 77.46$\pm$0.90            \\
Cat/Deer    & 100           & 61.58$\pm$2.07    & \textbf{66.47$\pm$2.04}   & 63.06$\pm$7.41    & 64.06$\pm$7.48            & 60.03$\pm$7.95            \\
            & 1000          & 70.94$\pm$1.49    & \textbf{71.36$\pm$1.61}   & 70.50$\pm$1.05     & 71.22$\pm$1.96            & 70.46$\pm$1.29            \\
            & 10000         & 77.30$\pm$0.85     & 77.73$\pm$0.69            & 79.96$\pm$0.94    & \textbf{81.42$\pm$0.83}   & 78.76$\pm$0.94            \\
Deer/Dog    & 100           & 61.18$\pm$1.79    & \textbf{65.12$\pm$1.39}   & 59.63$\pm$4.08    & 62.92$\pm$2.77            & 61.31$\pm$4.29            \\
            & 1000          & 68.91$\pm$1.24    & 70.04$\pm$1.09            & 70.52$\pm$2.15    & \textbf{71.78$\pm$3.55}   & 71.56$\pm$1.38            \\
            & 10000         & 76.33$\pm$0.71    & 77.66$\pm$0.64            & 82.61$\pm$1.00    & 84.46$\pm$0.96            & \textbf{84.85$\pm$0.69}   \\

\bottomrule
\end{tabular}
\end{sc}
\end{small}
\end{center}
\vskip -0.1in
\end{table*}

\subsection{Experiment of Empirical Kernels on CIFAR-10}
\label{cifar}

We empirically compare the CNTK and the WCNTK on  binary-classification on CIFAR-10 dataset, a widely used image dataset and contains 60,000 $32 \times 32$ colored images in 10 different classes. For each class, there are 5,000 training samples and 1,000 test samples. We test the performance of Vanilla CNN and the corresponding kernels (CNTK and WCNTK) on different subsets of CIFAR-10. 
Samples of different sizes ($10^2$, $10^3$, and $10^4$) are randomly selected for training, and we use the test set for evaluation. 

In this experiment, a five-layer Vanilla CNN, which consists of four convolutional layers and a fully connected layer, is used. The number of channels is 64, and the activation function is $tanh$. A fully connected layer with one neuron is used as the output. The neural network output is multiplied by a small constant in order that the second term of final prediction of the empirical CNTK and CWNTK (\ref{finitewidth}) becomes negligible without changing the descent direction.
For WNTK, layer-wise weights are learned by Algorithm~\ref{alg:algo1} with the following setting: initialized layer-wise weights $a^{L}_1 = 0.25$, $ a^{L}_i = 1\; (\forall i > 1)$, learning rate $\eta_w = 0.01$, and train-validation ratio $r = 0.2$. 
The regularization term in kernel regression is 1.

\textbf{Remark.} The selection of $a^{L}_1 = 0.25$ and $ a^{L}_i = 1\; (\forall i > 1)$ is motivated by the  fact that when applying Algorithm~\ref{alg:algo1}, the first layer's weight usually drops significantly to nearly 0. To speed up the convergence, we heuristically give a small initial value.


Each experiment is repeated ten times, and we report the mean, together with the standard deviation, in Table~\ref{table1}, where (W)NTK\_pur stands for (W)CNTK of CNN with randomly initialized parameters and (W)NTK\_ini for (W)CNTK of CNN with pretrained parameters. From Table~\ref{table1}, we could observe that:

\begin{itemize}
    \item The WNTK indeed significantly improves the performance of the NTK, by averagely $2.14\%$ for random initialization and $1.65\%$ for pretrained parameters. 
    
    \item Pretraining is helpful for both empirical NTKs and WNTKs,
    with which, both empirical NTKs and WNTKs can get a presentable improvement about $1\%$ to $2\%$.
    
    \item Empirical WNTKs could be competitive to or even better than the used NNs, especially if the pretraining is used and when the data size is not too large. 
    The used NNs are not wide enough and the weights of WNTKs can also be assigned element-wisely. These possibilities of amelioration make it promising to further improve the performance of the WNTK. 
\end{itemize}

\subsection{Experiment of Analytic Kernels on UCI}
\label{uci}

Next we consider tasks on UCI Machine Learning Repository. The analytic NTK and WNTK are evaluated on these non-image datasets, compared with a wide NN, simulating the corresponding infinite-width NN.
The selected task name, together with the size (S.), the number of features (F.), and the number of classes (F.) are shown in Table \ref{table2}. Different to the previous experiments, UCI data are non-image and we test the analytic WNTK of infinite-width NN, instead of an empirical one. 

Following the same setting of (\ref{wntkstability1}), the WNTK in this experiment is analytically generated from a five-layer fully connected NN with $ReLU$ activation under the infinite-width assumption. In pre-processing, all training data are standardized. The initial NN parameters are randomly given with unit parameter variance. The regularization term in kernel regression is set to be $0.1$. For comparison, we also train wide five-layer fully connected NNs, where the number of the channels is set to 1024 for both the input layer and every hidden layer, in order to simulate NNs of infinite width. 



In Table \ref{table2}, we report the accuracy calculated by 5-fold cross-validation. The WNTK improves accuracy from the NTK by $2\%$ to $3\%$. Such improvement reflects the fact that, under the infinite width assumption, the analytical WNTK is also better in generalization ability than the analytical NTK. Note that both WNTKs and NTKs are much better than fully connected NNs on small datasets. Such advantage of NN-induced kernels when facing small datasets has also been demonstrated by Arora et al. \yrcite{arora2019harnessing}.



\begin{table}[t]
\caption{Classification accuracy on 10 UCI datasets.}
\label{table2}
\vskip 0.05in
\begin{center}
\begin{small}
\begin{sc}
\begin{tabular}{ccccc}
\toprule
dataset         & s./f./c.       & NTK   & WNTK  & NN    \\
\midrule
tissue          & (106,9,6)     & 69.23 & 72.12 & 28.52 \\
echo            & (131,10,2)    & 81.82 & 84.09 & 75.78 \\
planning        & (182,12,2)    & 65.55 & 71.11 & 60.20  \\
cancer          & (198,33,2)    & 81.63 & 84.69 & 76.81 \\
heart           & (294,12,2)    & 83.56 & 85.62 & 78.41 \\
survival        & (306,3,2)     & 70.72 & 73.68 & 66.82 \\
tumor           & (330,17,15)   & 49.09 & 51.83 & 15.85 \\
ilpd            & (583,9,2)     & 71.23 & 73.29 & 66.65 \\
credit          & (690,14,2)    & 60.61 & 64.83 & 80.80  \\
contrac         & (1473,9,3)    & 48.37 & 52.51 & 39.99 \\

\bottomrule
\end{tabular}
\end{sc}
\end{small}
\end{center}
\vskip -0.1in
\end{table}

\section{Conclusion}
\label{conclusion}
Motivated by the fact that the Neural Tangent Kernel is restricted to describe NN dynamics with gradient descent, this paper proposes the Weighted Neural Tangent Kernel as a generalized and improved version of the NTK that can simulate wide NN dynamics with different optimizers. With rigorous discussion, we prove that the WNTK is also stable at initialization and during training. A sufficiently wide net fully trained by adjusted gradient descent is proved to be a linearized model and equivalent to the WNTK regression predictor. Hence, the WNTK possesses all properties of the NTK, and we believe that the WNTK is a more powerful tool to understand the correspondence between wide NNs and kernels. With developing a weight learning algorithm, we also verify the WNTK's superiority over the NTK.

\bibliography{paper}
\bibliographystyle{icml2021}

\clearpage

\appendix
\section{Appendix}
\subsection{Notations}
\begin{itemize}
\item $\otimes$ stands for the Kronecker product, and $\odot$ stands for the penetrating face product.
\item Without specific definition, $\dot \sigma$ represents the derivative of $\sigma$, and $\delta$ represents the Dirac delta function. $\delta_{kk'} = 0$ except for the case $k=k'$, $\delta_{kk} = 1$.
\item $||.||_{op}$ is the Operator Norm, and $||.||_{F}$ is the Frobenius Norm.
\end{itemize}

\subsection{Additional Experiment Settings}

In the CIFAR-10 experiment, pretraining is considered a trick to select a good initialization of $a$. It refers to training the NN with a small learning rate to a low accuracy threshold on the validation set before generating the empirical kernels. The idea of pretraining comes from the observation that the (W)NTK's performance is significantly improved if the corresponding NN is trained for some epochs. To ensure that the improvement of the NTK is not attributed to its corresponding NN, pretraining only attains a relatively low accuracy. For training sets of size $\leq 1,000$, the accuracy threshold is set to 60\%, and for others, the accuracy threshold is set to 70\%, where the NN is usually trained for only several epochs. In this experiment, the Adam optimizer is used for NN training. Each NN's accuracy is calculated using the parameters with the highest accuracy on the validation set trained in 200 epochs. Due to efficiency considerations, for all experiments, we do not use training tricks like batch normalization, dropout, data augmentation, Etc. 

We conduct four binary classification experiments on several datasets of CIFAR-10. These datasets are chosen because their NTK accuracy is not too high. The utilization of datasets with high NTK accuracy may influence the WNTK's performance improvement, as both methods are already close to the upper limit of classification accuracy. 

In our experiment, the WCNTK is calculated empirically. To  understand it better, we can find a detailed explanation of the analytic WCNTK in Arora et al. \yrcite{Arora2019OnEC}.

In the UCI experiment, the five-layer architecture inevitably reduces the performance of NNs. Such an architecture is chosen to better reflect the effect of weight terms in kernel estimators, as generally, kernel machines are more often used and have better performance on small datasets, compared to NNs. 

\subsection{Proof of Theorem 1.}

\textbf{Theorem 1 }\textit{\textbf{(Stability at Initialization).} Consider a network of depth L at initialization, with a Lipschitz nonlinearity $\sigma$. Suppose that, for every parameter $\theta_p$ in the $l$-th layer, its weight term in the WNTK follows a layer-wise distribution $\rho_l$. In the limit as the layer width $d_1,...,d_{L-1}$ tends to infinity sequentially, the layer-wise WNTK converges in probability to a deterministic limiting kernel}
\begin{align*}
\mathcal{A} \to \mathcal{A}^{(L)}_{\infty} \otimes Id_{d_L}.
\end{align*}
\textit{With $\mu_l$ referring to the expectation of all element-wise weights which follow the layer-wise distribution $\rho_l$, the scalar kernel $\mathcal{A}^{(L)}_{\infty}: {\mathbb{R}}^{d_0} \times {\mathbb{R}}^{d_0} \to \mathbb{R}$ is defined recursively by}
\begin{align*}
\mathcal{A}^{(1)}_{\infty}(x_i, x_j) =& \ \mu_1\Sigma^{(1)}(x_i, x_j), \\
\mathcal{A}^{(l)}_{\infty}(x_i, x_j) =& \ \mathcal{A}^{(l-1)}_{\infty}(x_i, x_j) \dot \Sigma^{(l)}(x_i, x_j)\\
& + \mu_{l}\Sigma^{(l)}(x_i, x_j),
\end{align*}
\textit{where}
\begin{align*}
\Sigma^{(1)}(x_i, x_j) &= \frac{1}{d_0}x^{\top}_ix_j,\\
\Sigma^{(l)}(x_i, x_j) &= \mathbb{E}_{f \sim \mathcal{N}(0, \Sigma^{(l-1)})}[\sigma(f(x_i))\sigma(f(x_j))],\\
\dot \Sigma^{(l)}(x_i, x_j) &= \mathbb{E}_{f \sim \mathcal{N}(0, \Sigma^{(l-1)})}[\dot\sigma(f(x_i))\dot\sigma(f(x_j))].
\end{align*}

\textbf{Proof.} The proof is by induction. In the case of depth $= 1$,
\begin{align*}
& \ \mathcal{A}^{(1)}_{kk'}(x_i, x_j) \\
=& \ \sum_{\mathrm{\theta_p \ in \ layer \ 1}} a_p [\partial_{\theta_{p}} f^{(1)}_{k}(\theta, x_i) \times \partial_{\theta_{p}} f^{(1)}_{k'}(\theta, x_j)]  \delta_{kk'}\\
\to& \ \mu_1\frac{1}{d_0}x^{T}_ix_j\delta_{kk'} \\
\to& \ \mu_1\Sigma^{(1)}(x_i, x_j)\delta_{kk'}\\
=& \ \mathcal{A}^{(1)}_{\infty}(x_i, x_j)\delta_{kk'}.
\end{align*}
So, 
\begin{align*}
\mathcal{A}^{(1)} \to \  \mathcal{A}^{(1)}_{\infty} \otimes Id_{d_1} .
\end{align*}
In the case of depth $= l+1$, by the induction hypothesis, we assume that 
\begin{align*}
\mathcal{A}^{(l)} \to \mathcal{A}^{(l)}_{\infty} \otimes Id_{d_l} ,
\end{align*}
or, equivalently, if we reduce the dimension of the last formula,
\begin{align*}
\mathcal{A}_{kk'}^{(l)}(x_i, x_j) \to \mathcal{A}^{(l)}_{\infty} (x_i, x_j) \delta_{kk'}.
\end{align*}
In the following, we will demonstrate $\mathcal{A}^{(l+1)} \to \mathcal{A}^{(l+1)}_{\infty} \otimes Id_{d_{l+1}}$.
\begin{align*}
& \mathcal{A}_{kk'}^{(l+1)} (x_i,x_j) \\
=& \ (\partial_{\theta} f_k^{(l+1)}(\theta, x_i))^{T} (a \odot \partial_{\theta} f_{k'}^{(l+1)}(\theta, x_j)) \delta_{kk'}.
\end{align*}
Let's divide the sum of all parameters into two parts: the sum of parameters in the first $l$ layer and the sum of parameters in the $(l+1)$ layer. In the first case, we have $\partial_{\theta} f^{(l+1)}_k(\theta, x_i)$ $= \frac{1}{\sqrt{d_l}}\sum^{d_l}_{m=1} \partial_{\theta}(f_k^{(l)}(\theta, x_i))\dot\sigma(f_k^{(l)}(\theta,x_i))W^{(l)}_{mk}$.

So, as $d_l \to \infty$, we can make the following approximation by the law of large number,
\begin{align*}
&(\partial_{\theta} f_k^{(l+1)}(\theta, x_i))^{T} (a \odot \partial_{\theta} f_{k'}^{(l+1)}(\theta, x_j)) \\
\to& \  \dot \Sigma^{(l+1)}(x_i, x_j) \sum_{m=1}^{d_l} \mathcal{A}^{(l)}_{kk'}(x_i,x_j) W^{(l)}_{mk}W^{(l)}_{mk'}\\
\to& \ \mathcal{A}^{(l)}_{\infty}(x_i,x_j) \dot \Sigma^{(l+1)}(x_i, x_j)\delta_{kk'}.
\end{align*}
In the second case, just similar to the case of depth $=1$, we can easily find out,
\begin{align*}
&(\partial_{\theta} f_k^{(l+1)}(\theta, x_i))^{T} (a \odot \partial_{\theta} f_{k'}^{(l+1)}(\theta, x_j))\\
\to & \ \mu_l\Sigma^{(l+1)}(x_i, x_j).
\end{align*}
Now, it is easy to conclude that,
\begin{align*}
& \mathcal{A}_{kk'}^{(l+1)}(x_i, x_j) \\
\to & \ (\mathcal{A}^{(l)}_{\infty}(x_i,x_j) \dot \Sigma^{(l+1)}(x_i, x_j) + \mu_l\Sigma^{(l+1)}(x_i, x_j))\delta_{kk'}\\
\to & \ \mathcal{A}^{(l+1)}_{\infty} (x_i, x_j) \delta_{kk'},
\end{align*}
or, equivalently,
\begin{align*}
\mathcal{A}^{(l+1)} \to \mathcal{A}^{(l+1)}_{\infty} \otimes Id_{d_{l+1}}.
\end{align*}
$\hfill\blacksquare$

\subsection{Proof of Theorem 2.}
\textbf{Assumptions [1-4]:} 
\begin{itemize}
\item The widths of the hidden layers are identical, i.e. $d_1 = ... = d_{L-1} = d$.
\item The analytic WNTK is full-rank. And we set $\eta_{\mathrm{critical}}$ $= 2(\lambda_{\min}+\lambda_{\max})^{-1} > 0 $. $\lambda_{\min}$ and $\lambda_{\max}$ are the smallest and the largest eigenvalues of the WNTK. $\lambda_{\max} \ge \lambda_{\min} > 0$.
\item The training set $(\mathcal{X},\mathcal{Y})$ is contained in some compact sets and $x_i \ne x_j$ for all $x_i,x_j \in \mathcal{X}$.
\item The activation function $\sigma$ satisfies: $|\sigma(0)|$, $||\dot\sigma||_{\infty}$, $\sup_{x_i \ne x_j}|\dot\sigma(x_i) - \dot\sigma(x_j)|/|x_i-x_j| < \infty $.
\end{itemize}

\begin{lemma}
\label{l1}
    \textbf{(Local Lipschitzness of the Jacobean).} \textit{There is a $K > 0$ such that for every $C > 0$, with high probability over random initialization (w.h.p.o.r.i.) the following holds:}
    \begin{align*}
    &\begin{cases}
    d^{-\frac{1}{2}}\frac{||J(\theta) - J(\theta')||_{F}}{||\theta - \theta'||_{2}} \leq K\\
    d^{-\frac{1}{2}}||J(\theta)||_{F}\leq K
    \end{cases}, \forall \theta, \theta' \in B(\theta_{0}, Cd^{-\frac{1}{2}}),
    \end{align*}
    \textit{where}
    \begin{align*}
    B(\theta_{0},R) :=\{\theta:||\theta - \theta_{0}||_2<R\} \textit{\ and \ } J(\theta) \in \mathbb{R}^{n\times P}.
    \end{align*}
\end{lemma}
Note that $d_L$ is assumed to be 1 in the following. The proof of Lemma \ref{l1} can be found in Lee's work \yrcite{lee2019wide}. The proof of Theorem \ref{t2} relies on the Local Lipschitzness of the Jacobean.

\textbf{Theorem 2 }\textit{\textbf{(Stability during Training).} Under Assumptions 1-4 and suppose that an  L-layer neural network $f$ with random initialization is trained by gradient decent with different learning rate $\eta a_p$ on different parameter $\theta_p$,  where $\eta=\frac{\eta_0}{d}$. 
For $\delta_0 >0$ and $\eta_0<\eta_{\mathrm{critical}}$, there exist $R_0 > 0$, $N \in \mathbb{N}$, and $K>1$ such that for every $d \ge N$, 
we have the following inequality with probability at least $(1-\delta_0)$:}
\begin{align*}
||\mathcal{A}_{0} - \mathcal{A}_{t}||_{F} \leq \frac{6K^{3}R_0a_{\max}^2}{\lambda_{\min}}d^{-\frac{1}{2}}, \forall t,
\end{align*}
\textit{where $\mathcal{A}_0$ and $\mathcal{A}_t$ are the WNTK at moment $0$ and $t$ under standard parameterization, and $a_{\max} = \max\{a_i| i = 1,...,P\}$.}

\textbf{Proof.} First, take $K$ and $D_0$ big enough that the Local Lipschitzness of the Jacobean holds with the probability $(1-\delta_0/3)$, $\forall d>D_0$. Let's assume that $\forall t, \theta_t \in B(\theta_{0}, Cd^{-\frac{1}{2}})$ with $C=\frac{3KR_0a_{\max}}{\lambda_{\min}}$.
According to the definition of WNTK and Lemma \ref{l1}, we have
\begin{align*}
& \ ||\mathcal{A}_{0} - \mathcal{A}_{t}||_{F} \\
=& \  \frac{1}{d}||J(\theta_0)(a \odot J(\theta_0)^T) - J(\theta_t)( a \odot J(\theta_t)^T)||_F\\
\leq& \ \frac{a_{\max}}{d}(||J(\theta_0)||_{op}||J(\theta_0)^T - J(\theta_t)^T||_{F}\\
& \ + ||J(\theta_t)||_{F}||J(\theta_t)^T - J(\theta_0)^T||_{op})\\
\leq& \ a_{\max}2K^2||\theta_0-\theta_t||_2\\
\leq& \ \frac{6K^{3}R_0a_{\max}^2}{\lambda_{\min}}d^{-\frac{1}{2}}.
\end{align*}
So, our target is now to prove by induction the following inequality with a known small probability:
\begin{align*}
||\theta_t-\theta_0||_2 \leq \frac{3KR_0a_{\max}}{\lambda_{\min}}d^{-\frac{1}{2}},\forall t.
\end{align*}
At initialization, by applying Lemma \ref{l1}, We define $g(\theta) = f(\theta,\mathcal{X}) - \mathcal{Y}$ and we have
\begin{align*}
||\theta_1-\theta_0||_2 \leq & \ \eta a_{\max} ||J(\theta_0)||_{op}||g(\theta_0)||_2 \\
\leq & \ \eta_0Ka_{\max}||g(\theta_0)||_2d^{-\frac{1}{2}}.
\end{align*}
Since $f$ converges in distribution to a mean zero Gaussian, there exists $R_0$ and $D_1$, such that for every $d>D_1$, with probability at least $(1-\delta/3)$ over random initialization,
\begin{align*}
||g(\theta_0)||_2 < R_0.
\end{align*}
According to Assumption 2, we have $\eta_0\leq \frac{2}{\lambda_{\min}+\lambda_{\max}}$. Hence, $\eta_0 \leq \frac{3}{\lambda_{\min}}$ and we can prove that,
\begin{align*}
||\theta_1-\theta_0||_2 \leq \frac{3KR_0a_{\max}}{\lambda_{\min}}d^{-\frac{1}{2}}.
\end{align*}
From now on, we will prove the part of iteration. We assume that $\forall t'\leq t, ||\theta_{t'}-\theta_0||_2 \leq \frac{3KR_0a_{\max}}{\lambda_{\min}}d^{-\frac{1}{2}}$. We have
\begin{align*}
||\theta_{t+1} - \theta_0||_2 &\leq \sum_{t' = 0}^{t} ||\theta_{t'+1}-\theta_{t'}||_2\\
& \leq \eta a_{\max} \sum_{t' = 0}^{t} ||J(\theta_t')||_{op}||g(\theta_{t'})||_2\\
& \leq \eta_0 Ka_{\max}d^{-\frac{1}{2}} \sum_{t' = 0}^{t}||g(\theta_{t'})||_2.
\end{align*}
Suppose we have $\forall t'\leq t,||g(\theta_{t'})||_2\leq (1-\frac{\eta_0\lambda_{\min}}{3})^{t'}R_{0}$. It can be deduced that $\sum_{t' = 0}^{t}||g(\theta_{t'})||_2 \leq \frac{3}{\eta_0\lambda_{min}}R_0$, and $||\theta_{t+1} - \theta_0||_2 \leq \frac{3KR_0a_{\max}}{\lambda_{\min}} d^{-\frac{1}{2}}$. So, in the rest of this proof, we will demonstrate the existence of the upper bound of $g(\theta_{t'}), \forall t'\leq t$. We will still finish this proof by induction.

At initialization, it has been discussed above that $g(\theta_{0})\leq R_0$. For the part of iteration, we assume that we already have $\forall t^{\star}\leq t',||g(\theta_{t^{\star}})||_2\leq (1-\frac{\eta_0\lambda_{\min}}{3})^{t^{\star}}R_{0}$. So,
\begin{align*}
||g(\theta_{t'+1})||_2 =& \ ||g(\theta_{t'+1}) - g(\theta_{t'}) + g(\theta_{t'})||_2\\
=& \ ||J(\theta_{t'})(\theta_{t'+1} - \theta_{t'}) + g(\theta_{t'})||_2\\
=& \ ||-\eta J(\theta'_{t'})(a\odot J(\theta_{t'})^T)g(\theta_{t'})+g(\theta_{t'})||_2\\
\leq& \ ||1 - \eta J(\theta'_{t'})(a\odot J(\theta_{t'})^T)||_{op}||g(\theta_{t'})||_2\\
\leq& \ ||1 - \eta J(\theta'_{t'})(a\odot J(\theta_{t'})^T)||_{op}\\
& \ (1-\frac{\eta_0\lambda_{\min}}{3})^{t'}R_0.
\end{align*}
where where $\theta'_{t'}$ is some linear interpolation between $\theta_{t'}$ and $\theta_{t'+1}$. It remains to show with a known small probability
\begin{align*}
||1 - \eta J(\theta'_{t'})(a \odot J(\theta_{t'})^T)||_{op} \leq 1-\frac{\eta_0\lambda_{min}}{3}.
\end{align*}
We have,
\begin{align*}
& \ ||1 - \eta J(\theta'_{t'})(a \odot J(\theta_{t'})^T)||_{op}\\
\leq& \  ||1 - \eta_0 \mathcal{A}_{\infty}^{(L)}||_{op} + \eta_0||\mathcal{A}_{\infty}^{(L)} - \mathcal{A}_0||_{op} \\
& \ +\eta||J(\theta_0)(a\odot J(\theta_0)^T) - J(\theta'_{t'})(a\odot J(\theta_{t'})^T)||_{op}\\
\leq& \  ||1 - \eta_0 \mathcal{A}_{\infty}^{(L)}||_{op} + \eta_0||\mathcal{A}_{\infty}^{(L)} - \mathcal{A}_0||_{op}\\
& \  +a_{\max}\eta_0K^2(||\theta_{t'}-\theta_0||_{2} + ||\theta'_{t'}-\theta_0||_{2}).
\end{align*}
Assumption 2 implies 
\begin{equation}
\label{t2.1}
||1 - \eta_0 \mathcal{A}_{\infty}^{(L)}||_{op} \leq 1-\eta_0\lambda_{\min}.
\end{equation}
As we have proved the convergence of $\mathcal{A}_0$ in Theorem \ref{t1}, one can find $D_2$ independent of $t'$, such that the following event holds with probability $(1-\delta_0/3)$,
\begin{equation}
\label{t2.2}
||\mathcal{A}_{\infty}^{(L)} - \mathcal{A}_0||_{op}\leq \frac{\lambda_{\min}}{3}.
\end{equation}
With (\ref{t2.1}), (\ref{t2.2}) and Lemma \ref{l1}, we have
\begin{align*}
& \ ||1 - \eta J(\theta'_{t'})(a \odot J(\theta_{t'})^T)||_{op}\\
\leq& \ 1 -\eta_0\lambda_{\min} + \frac{\eta_0\lambda_{\min}}{3} + 2a_{\max}^2\eta_0K^2\frac{3KR_0}{\sqrt{d}\lambda_{\min}}\\
\leq& \ 1 - \frac{\eta_0\lambda_{\min}}{3},
\end{align*}
under the constraint that,
\begin{align*}
d \geq (\frac{18K^3R_0a_{\max}^2}{\lambda_{\min}^2})^2 = D_3.
\end{align*}
Therefore, we only need to set $N = \max\{D_0, D_1, D_2, D_3\}$ and we can finally conclude that,
\begin{align*}
||\mathcal{A}_{t} - \mathcal{A}_{0}||_{F} \leq \frac{6K^{3}R_0a_{\max}^2}{\lambda_{\min}}d^{-\frac{1}{2}} = \mathcal{O}(d^{-\frac{1}{2}}), \forall t,
\end{align*}
with probability $p \geq (1-\delta_0/3)^3 \geq (1-\delta_0)$. $\hfill\blacksquare$

\subsection{Proof of Theorem 3.}

\textbf{Theorem 3 }\textit{\textbf{(WNTK Lazy Regime).} Under Assumptions 1-4, applying gradient descent with different learning rate $\eta a_p$ on different parameter $\theta_p$, for every $x \in  {\mathbb{R}}^{d_0}$ satisfying $||x||_2 \leq 1$, for all $t\ge0$, as $ d \to \infty$, with probability arbitrarily close to 1 over random initialization, we have}
\begin{align*} 
|| f(\theta_t, \mathcal{X}) - f^{\mathrm{lin}}_t(\mathcal{X})||_{2} \to 0,\\
|| f(\theta_t, x) - f^{\mathrm{lin}}_t(x)||_{2} \to 0.
\end{align*}

\textbf{Proof.} This proof follows a similar proof in Lee et al. \yrcite{lee2019wide}. Via continuous time gradient descent, the evolution of the parameters and the output function of the neural network in this theorem can be written as follows.
\begin{align*}
\dot\theta_t &= - \eta (a \odot \nabla_{\theta} f(\theta_t,\mathcal{X}))^{T} \nabla_{f(\theta_t,\mathcal{X})}\mathcal{L},\\
\dot f(\theta_t,x) &= - \eta \mathcal{A}_t(x,\mathcal{X})\nabla_{f(\theta_t,\mathcal{X})}\mathcal{L}.
\end{align*}
Correspondingly, the dynamics of gradient flow using the linearized model are governed by,
\begin{align*}
\dot \theta_t &= - \eta (a \odot \nabla_{\theta} f_0(\mathcal{X}))^{T} \nabla_{f^{\mathrm{lin}}_t(\mathcal{X})}\mathcal{L},\\
\dot f^{\mathrm{lin}}_t(x) &= - \eta \mathcal{A}_0(x,\mathcal{X})\nabla_{f^{\mathrm{lin}}_t(\mathcal{X})}\mathcal{L}.
\end{align*}
We define $g^{\mathrm{lin}}(t) = f^{\mathrm{lin}}_t(\mathcal{X}) - \mathcal{Y}$ and $g(t) = f(\theta_t,\mathcal{X}) - \mathcal{Y}$. We begin our proof by the following equation:
\begin{align*}
    &\frac{d}{dt} (\exp(\eta\mathcal{A}_0t)(g^{\mathrm{lin}}(t) - g(t)) \\
    =& \eta(\exp(\eta\mathcal{A}_0t)(\mathcal{A}_t - \mathcal{A}_0)g(t)).
\end{align*}
Integrating both sides and using the fact $g^{\mathrm{lin}}(0) = g(0)$,
\begin{align*}
    & (g^{\mathrm{lin}}(t)-g(t)) = \\
    & - \int_0^t \eta(\exp(\eta \mathcal{A}_0 (s-t))( \mathcal{A}_s - \mathcal{A}_0)(g^{\mathrm{lin}}(s)-g(s)))ds\\
    & + \int_0^t \eta(\exp(\eta \mathcal{A}_0 (s-t))( \mathcal{A}_s - \mathcal{A}_0)g^{\mathrm{lin}}(s))ds.
\end{align*}
Taking the norm gives
\begin{align*}
    & ||g^{\mathrm{lin}}(t)-g(t)||_2 \leq \\
    & \eta \int_0^t ||\exp^{\eta \mathcal{A}_0 (s-t)}||_{op}||\mathcal{A}_s - \mathcal{A}_0||_{op}||g^{\mathrm{lin}}(s)-g(s)||_2ds\\
    & + \eta \int_0^t ||\exp^{\eta \mathcal{A}_0 (s-t)}||_{op}||\mathcal{A}_s - \mathcal{A}_0||_{op}||g^{\mathrm{lin}}(s)||_2ds.
\end{align*}
Let $\lambda_0 >0$ be the smallest eigenvalue of $\mathcal{A}_0$.
\begin{align*}
    & ||g^{\mathrm{lin}}(t)-g(t)||_2 \leq \\
    & \eta \int_0^t \exp^{\eta \lambda_0 (s-t)}||\mathcal{A}_s - \mathcal{A}_0||_{op}||g^{\mathrm{lin}}(s)-g(s)||_2ds\\
    & + \eta \int_0^t \exp^{\eta \lambda_0 (s-t)}||\mathcal{A}_s - \mathcal{A}_0||_{op}||g^{\mathrm{lin}}(s)||_2ds.
\end{align*}
We conduct the following variable replacement. Let
\begin{align*}
    u(t) &= \exp^{\eta \lambda_0 t}||g^{\mathrm{lin}}(s)-g(s)||_2,\\
    \alpha(t) &= \eta \int_0^t \exp^{\eta \lambda_0 (s)}||\mathcal{A}_s - \mathcal{A}_0||_{op}||g^{\mathrm{lin}}(s)||_2ds,\\
    \beta(t) &= \eta ||\mathcal{A}_t - \mathcal{A}_0||_{op} .
\end{align*}
The above inequality can be rewritten as
\begin{align*}
    u(t) \leq \alpha(t) + \int_{0}^{t}\beta{s}u(s)ds.
\end{align*}
By applying  an integral form of the Gronwall's inequality \cite{dragomir2003some}, we can get:
\begin{align*}
    u(t) \leq \alpha(t)\exp(\int_{0}^{t}\beta{s}ds).
\end{align*}
According to the dynamics of the linearized network, we have 
\begin{align*}
    ||g^{\mathrm{lin}}(t) - g(t)||_2 \leq \exp^{-\lambda_0 \eta t}||g^{\mathrm{lin}}(0)||_2.
\end{align*}
Then
\begin{equation}
    \label{T3.1}
    \begin{aligned}
    & ||g^{\mathrm{lin}}(t)-g(t)||_2 \leq \\
    & \exp^{-\lambda_0 \eta t} ||g^{\mathrm{lin}}(0)||_2 \int_0^t \beta(s)ds \exp(\int_0^t \beta(s)ds).
    \end{aligned}
\end{equation}
Let $\sigma_t = \sup_{0\leq s\leq t}||\mathcal{A}_s - \mathcal{A}_0||_{op}$. According to Theorem 2,
\begin{equation}
    \label{T3.2}
    \begin{aligned}
    \forall t, \sigma_t = O(d^{-\frac{1}{2}}) \to 0.
    \end{aligned}
\end{equation}
Thus there exists $P(t)$ a polynomial, such that
\begin{align*}
    ||g^{\mathrm{lin}}(t)-g(t)||_2 \leq (\eta t \sigma_t \exp^{-\lambda\eta t + \sigma_t \eta t})||g^{\mathrm{lin}}(0)||_2 P(t) .
\end{align*}
Therefore, 
\begin{align*}
    &|| f(\theta_t, \mathcal{X}) - f^{\mathrm{lin}}_t(\mathcal{X})||_{2} \\
    = &||g(t)-g^{\mathrm{lin}}(t)||_2 \leq O(d^{-\frac{1}{2}}) \to 0.
\end{align*}
Similarly, we can prove the convergence on a test sample. Let $y$ be the true label and We define $g^{\mathrm{lin}}(t,x) = f^{\mathrm{lin}}_t(x) - y$ and $g(t,x) = f(\theta_t,x) - y$. We have
\begin{align*}
    & \frac{d}{dt}(g^{\mathrm{lin}}(t,x)-g(t,x)) \\
   = & - \eta (\mathcal{A}_0(x,\mathcal{X}) - \mathcal{A}_t(x,\mathcal{X}))g^{\mathrm{lin}}(t) \\
    & + \eta \mathcal{A}_t(x,\mathcal{X}) (g(t) - g^{\mathrm{lin}}(t)).
\end{align*}
We conduct the same integration over $[0,t]$ and take the norm.
\begin{align*}
    & ||g^{\mathrm{lin}}(t,x)-g(t,x)||_2\\
    \leq & \eta \int_0^t \exp^{-\eta \lambda_0 s}||\mathcal{A}_s(x,\mathcal{X}) - \mathcal{A}_0(x,\mathcal{X})||_{2}ds\\
    & + \eta \int_0^t (||\mathcal{A}_0(x,\mathcal{X})||_{2} + ||\mathcal{A}_s(x,\mathcal{X}) - \mathcal{A}_0(x,\mathcal{X})||_{2})\\
    & ||g(s) - g^{\mathrm{lin}}(s)||_2ds.
\end{align*}
Lemma \ref{l1} can imply, for all $t$,
\begin{align*}
||\mathcal{A}_s(x,\mathcal{X}) - \mathcal{A}_0(x,\mathcal{X})||_{2} = O(d^{-\frac{1}{2}}).
\end{align*}
This implies that the first term in the right side of the inequality $= O(d^{-\frac{1}{2}})$. In terms of the the second term, we can find its upper-bound by exactly the same way as in the proof on the training set. By applying (\ref{T3.1}) and (\ref{T3.2}), we can prove that the second term $\leq ||\mathcal{A}_0(x,\mathcal{X})||_{2} \int_0^t (\eta s \sigma_s \exp^{-\eta \lambda_0 s + \sigma_s\eta s})||g^{\mathrm{lin}}(0)||_2 dt P(t) = O(d^{-\frac{1}{2}})$. So, we end the whole proof by
\begin{align*}
    & \ || f(\theta_t, x) - f^{\mathrm{lin}}_t(x)||_{2}\\
    = & \ ||g(t,x) - g^{\mathrm{lin}}(t,x)||_2 \leq O(d^{-\frac{1}{2}}) \to 0.
\end{align*}
$\hfill\blacksquare$

\end{document}